\documentclass{sig-alternate-05-2015}
\usepackage{tablefootnote}

\begin{document}

\CopyrightYear{2017}
\setcopyright{rightsretained}
\conferenceinfo{SIGCSE '17}{March 08-11, 2017, Seattle, WA, USA} \isbn{978-1-4503-4698-6/17/03} \doi{http://dx.doi.org/10.1145/3017680.3022464}
\clubpenalty=10000 
\widowpenalty = 10000

\numberofauthors{1} 
%
\author{
%
%
\alignauthor
Haoze Wu\\
        \affaddr{Department of Mathematics and Computer Science}\\
       \affaddr{Davidson College}\\
       \affaddr{Davidson, NC, USA}\\
       \email{anwu@davidson.edu}
\alignauthor
Raghuram Ramanujan\\
       \affaddr{Davidson College}\\
       \affaddr{P.O. Box 5996, 209 Ridge Rd}\\
       \affaddr{Davidson, NC, USA}\\
       \email{raramanujan@davidson.edu}
}
\additionalauthors{Additional authors: John Smith (The Th{\o}rv{\"a}ld Group,
email: {\texttt{jsmith@affiliation.org}}) and Julius P.~Kumquat
(The Kumquat Consortium, email: {\texttt{jpkumquat@consortium.net}}).}
\date{30 July 1999}

\title{Improving SAT-solving with Machine Learning}

\maketitle

\begin{abstract}
In this project, we aimed to improve the runtime of Minisat, a Conflict-Driven Clause Learning (CDCL) solver that solves the Propositional Boolean Satisfiability (SAT) problem. We first used a logistic regression model to predict the satisfiability of propositional boolean formulae after fixing the values of a certain fraction of the variables in each formula. We then applied the logistic model and added a preprocessing period to Minisat to determine the preferable initial value (either true or false) of each boolean variable using a Monte-Carlo approach. Concretely, for each Monte-Carlo trial, we fixed the values of a certain ratio of randomly selected variables, and calculated the confidence that the resulting sub-formula is satisfiable with our logistic regression model. The initial value of each variable was set based on the mean confidence scores of the trials that started from the literals of that variable. We were particularly interested in setting the initial values of the backbone variables correctly, which are variables that have the same value in all solutions of a SAT formula. Our Monte-Carlo method was able to set 78$\%$ of the backbones correctly. Excluding the preprocessing time, compared with the default setting of Minisat, the runtime of Minisat for satisfiable formulae decreased by $23\%$. However, our method did not outperform vanilla Minisat in runtime, as the decrease in the conflicts was outweighed by the long runtime of the preprocessing period.

\end{abstract}

\keywords{3SAT; Satisfiability;  Logistic; Monte-Carlo; Machine Learning}

\section{Problem and Motivation}
In the Propositional Boolean Satisfiability (SAT) problem, one is given a Boolean formula, i.e., an expression that consists of Boolean variables connected by the fundamental Boolean operators "and", "or" and "not". One is then tasked with determining whether there is an assignment of true/false values to the variables such that the overall formula evaluates to true. For instance, $P \wedge (Q \vee \neg R)$, is a Boolean formula with three Boolean variables. This formula evaluates to true when $P$, $Q$, and $R$ are all set to true. Most state-of-the-art complete SAT solvers use the Conflict-Driven Clause Learning (CDCL) algorithm. Though able to solve large industrial formulae, the CDCL algorithm cannot efficiently solve random formulae of even moderate size (300-500 variables). Our goal is to improve the runtime of Minisat, a CDCL SAT solver, on random Boolean formulae, with machine learning. 

\section{Background and Related Work}

The SAT problem is one of the most studied NP-complete problems because of its theoretical significance and practical applications\cite{zhang2001efficient}.  Our work is inspired by the works of Xu, et al. who successfully applied machine learning to SAT problems. In \cite{satzilla}, they used a machine learning algorithm to create an ensemble SAT-solver, SATzilla, which uses several SAT-solvers as subroutines and makes a per-instance decision on which solver to pick for a given input formula.  In another work \cite{sat_pred}, Xu, et al used machine learning to predict the satisfiability of hard Boolean formulae and obtained 70\% accuracy. 
In our project, instead of using machine learning to optimize the selection of SAT-solvers, we aimed to exploit the learnability of satisfiability to improve the performance of a particular solver, Minisat \cite{sat}. Minisat uses the CDCL algorithm, which selects a variable from the formula to "branch on", fixes its value to either true or false, simplifies the formula with this assignment, and then recursively solves the rest of the formula. When a conflict (i.e., a contradiction) is detected at a certain branching level, the algorithm backtracks. Though Minisat uses a heuristic to optimize the selection of the branching variable\cite{sat}, it does not make intelligent choices when assigning values to these variables. We aim to use machine learning to set the default branching values of variables before running the CDCL algorithm, in order to decrease the number of conflicts, and thus decrease the runtime of Minisat.

\section{Approach and Uniqueness}
Our project relied on the hypothesis that if at each branching point, the branching variable is set to the value that is more likely to lead to a solution, then a solution would be found relatively quickly. 

We first trained a logistic regression model to predict the satisfiability of random Boolean formulae represented in conjunctive normal form(CNF). A CNF formula is composed of a conjunction of clauses, where each clause is the disjunction of variables (or their negations). A 3-CNF formula is one in which each clause contains exactly three variables (or their negations) --- for example, $(x_1 \lor x_2 \lor x_4) \wedge (\neg x_2 \lor x_3 \lor \neg x_4) \wedge (x_1 \lor \neg x_3 \lor \neg x_4)$. There is a well-known polynomial-time procedure for converting arbitrary SAT formulae into 3-CNF form\cite{convert}, so this restriction in clause length does not compromise generality.

We created the training data for our learning model by first generating random 3-CNF formulae with 300 variables. Since we planned to apply our fitted model to make predictions about the formulae that are generated by the CDCL process, we generated additional instances by randomly fixing $n\%$ of the variables in these formulae, for varying values of $n$. We extracted 10 features from each of these formulae to create individual training examples --- these features are listed in Table 1. The selection of the features was inspired by \cite{sat_pred}. The target variable is a binary value that indicated whether the formula in question was satisfiable. Building this regression model constitutes the offline phase of our new solver.

\begin{table}
\begin{center}
\begin{tabular}{ |c|c|c| } 
 \hline
1 & Clause to variable ratio  \\  \hline
2 & Fraction of binary clauses  \\  \hline
3 & Fraction of horn clauses  \\  \hline
4 & POSNEG\_ratio\_var\_max  \\ \hline
5 & POSNEG\_ratio\_var\_min \\ \hline
6 & POSNEG\_ratio\_var\_mean \\ \hline
7 & POSNEG\_ratio\_var\_std  \\  \hline
8 & POSNEG\_ratio\_var\_variation \\ \hline
9 & LPSLACK\_mean  \\ \hline
10 & LPSLACK\_coeff\_variation\\ \hline
\end{tabular}
\caption{We refer the reader to \cite{sat_pred} for the definitions of features 4-10}
\end{center}
\end{table}

In the online phase of the solver, we added a preprocessing period to Minisat to determine the preferred initial value of each Boolean variable with a Monte-Carlo approach. Concretely, to determine the preferred initial value of a variable $v$, we first set $v$ to true and conducted a Monte-Carlo simulation starting from the remaining (simplified) formula. In each Monte-Carlo trial, we randomly fixed the values of $n\%$ of the variables, and calculated the likelihood that the resulting formula was satisfiable, with our logistic regression model. A mean likelihood score was computed by averaging the scored from the outcomes of many trials. We repeated the same operations with $v$ set to false. We then set $v$ to the value (i.e., true or false) that suggested a higher chance of leading to a satisfiable solution, based on the outcome of the Monte-Carlo simulations. We were particularly interested to gauge the effectiveness of our model in correctly setting the values of so-called "backbone variables" --- those variables that have the same value in all solutions of a SAT formula \cite{backbone}.

We evaluated the performance of the preprocessing period by the ratio of backbones it set correctly, the number of conflicts the subsequent CDCL run yielded, and the actual runtime of Minisat. Since our method was geared towards finding solutions quickly, we focused our attention on satisfiable Boolean formulae. 

\section{Results and Contributions}
As shown by Table 2, our regression model achieved an accuracy of $70\%$ in predicting satisfiability when $n$ was $0\%$, and an accuracy of $78\%$ when $n$ was set to $4\%$. The best performance of the Monte-Carlo method was obtained when $n$ was set to $4\%$. On average, it set $78\%$ of the backbones correctly. 

\begin{table}
\begin{center}
\begin{tabular}{ |c|c|c| } 
 \hline
$n$ & SAT Prediction Score  & Backbone setting Score \\ \hline
4\% & 77.94\% & 77.56\%\\  \hline
2\% & 68.56\% & 77.56\%\\  \hline
0\% & 70.38\% & 68.00\% \\ \hline
\end{tabular}
\caption{The performance of SAT prediction and backbone setting with different depths of Monte-Carlo trails}
\end{center}
\end{table}

Compared with the default setting of Minisat (which always sets the branching variable to false), our method yielded an average decrease in conflicts of $23\%$ and outperformed default Minisat in terms of conflicts in $55\%$ of the test cases. However, it did not outperform vanilla Minisat in runtime, as the decrease in the conflicts was outweighed by the long runtime of the preprocessing period.

\bibliographystyle{abbrv}
\bibliography{sigproc}  
%
%

\end{document}